\documentclass{article}

\usepackage[preprint]{neurips_2023}
\usepackage{amsmath}
\usepackage{multirow}
\usepackage{graphicx}
\usepackage{epsfig}
\usepackage{subfig}
\usepackage{float}
\usepackage{multirow}
\usepackage{appendix}
\usepackage{color}
\usepackage{times}
\usepackage{helvet}
\usepackage{courier}
\usepackage{multirow}
\usepackage{bm}
\usepackage[american]{babel}
\usepackage{enumitem}
\usepackage{caption}
\usepackage{diagbox}
\setcitestyle{square,comma,sort,numbers}
\usepackage[utf8]{inputenc} 
\usepackage[T1]{fontenc}    
\usepackage{hyperref}       
\usepackage{url}            
\usepackage{booktabs}       
\usepackage{amsfonts}       
\usepackage{nicefrac}       
\usepackage{microtype}      
\usepackage{xcolor}         
\usepackage{algorithm}
\usepackage{algpseudocode}

\title{NODI: Out-Of-Distribution Detection with Noise from Diffusion}

\author{%
  Jingqiu Zhou\\
  Chinese University of HongKong\\
  \texttt{} \\
  \And
   Aojun Zhou \\
  Chinese University of HongKong \\
  \texttt{} \\
  \AND
  Hongsheng Li \\
  Chinese University of HongKong\\
  \texttt{} \\
}

\begin{document}

\maketitle

\begin{abstract}
Out-of-distribution (OOD) detection is a crucial part of deploying machine learning models safely. It has been extensively studied with a plethora of methods developed in the literature. This problem is tackled with an OOD score computation, however, previous methods compute the OOD scores with limited usage of the in-distribution dataset. For instance, the OOD scores are computed with information from a small portion of the in-distribution data. Furthermore, these methods encode images with a neural image encoder. The robustness of these methods is rarely checked with respect to image encoders of different training methods and architectures. In this work, we introduce the diffusion process into the OOD task. The diffusion model integrates information on the whole training set into the predicted noise vectors. What's more, we deduce a closed-form solution for the noise vector (stable point). Then the noise vector is converted into our OOD score, we test both the deep model predicted noise vector and the closed-form noise vector on the OOD benchmarks \cite{openood}. Our method outperforms previous OOD methods across all types of image encoders (Table. \ref{main}). A $3.5\%$ performance gain is achieved with the MAE-based image encoder. Moreover, we studied the robustness of OOD methods by applying different types of image encoders. Some OOD methods failed to generalize well when switching image encoders from ResNet to Vision Transformers, our method performs exhibits good robustness with all the image encoders.

\end{abstract}

\section{Introduction}
"Is the model making a faithful prediction?" is always a vital question to be asked when deploying a machine learning model. The existing machine learning models are usually trained with a closed dataset, thus they implicitly assumed the close-world assumption at test time. However, this close-world assumption can be easily violated in real-world applications \cite{openworld}. The deployed machine learning model should not only give high confidence to testing samples in the training distribution but also be able to detect irrelevant samples. This leads to the important task of out-of-distribution detection (OOD).

Previous works \cite{msp, odin, lee2017training, lee2018simple, yu2019unsupervised,gradnorm_ood, openood} have provided plenty of benchmarks and achieved remarkable results in this field. The existing OOD methods all focus on computing OOD scores, and the way of generating OOD scores can be categorized \cite{ood_survey} into three types: classification-based, density-based, and distance-based. For classification-based methods \cite{msp, odin,godin, mls, zhou2023improving}, they rely on the confidence scores of the classification models. They assume that samples with high classification scores are in-distribution ones. The density-based methods \cite{abati2019latent,pidhorskyi2018generative, zong2018deep} usually model the distribution with probabilistic models. For the distance-based methods \cite{lee2018simple, ren2021simple,sun2022out}, they compute the distance between samples in the feature space and predict OOD scores based on the distances. Despite the great success achieved by these methods, they share some common drawbacks: 1). All the OOD methods focus on manipulation of the in-distribution data because they lack proper tools for generating realistic OOD data; 2). Only a small portion of data in the training set is used when computing OOD scores. The influences of data far away from the testing sample are usually ignored.

 To address the above-mentioned drawbacks, we explore the possibility of solving the OOD problem with a diffusion model. For the first drawback, this task usually lacks a good typical set of OOD samples. Although one can collect OOD samples and use them in the training process \cite{lee2017training}, it is always possible to find samples outside the distribution of collected OOD samples. These methods suffer from performance drops on unseen OOD samples. The diffusion model provides a workaround to this issue because it generates OOD data on-the-fly during the training process. Although the separation hyperplane between in-distribution and out-of-distribution data is unknown, the diffusion model can be used as a tool for estimating the distance between the out-of-distribution and in-distribution samples. For the second drawback, unlike the previous distance-based method \cite{sun2022out} which focuses on determining OOD scores with the top-$k$ closest in-distribution data points, the diffusion model integrated all the information in the predicted noise vector of our training set. To better understand the noise vectors predicted by our diffusion model, we deduce a closed-form noise vector formula. The diffusion model outputs this closed-form noise vector if it is properly trained. We evaluate the OOD benchmark provided by \cite{openood} with both the diffusion model and the closed-form noise vector. Furthermore, we investigated the robustness of our OOD methods with different image neural encoders, including both supervised and unsupervised ones.
 
 However, there is a remaining obstacle that needs to be addressed for the full application of the diffusion model.

Mapping the noise predicted by diffusion model \cite{ddpm,ddim} to out-of-distribution score is a non-trivial process. One can naively use the noise vector magnitude as the out-of-distribution score, however, this strategy assumes that the magnitude of the noise vector positively correlated to the out-of-distribution scores almost everywhere in the latent space which has difficulty holding. For data points far away from the training set in terms of $l_2$ distance, they are rarely diffused to and resulting in a noise vector with a large length. These data points can be both in-distribution or out-of-distribution. To address this issue, we confine all the in-distribution samples on the sphere's surface with a predefined radius. In this way, testing features with large magnitudes can be identified as OOD samples. A new issue appears that we can not naively apply the diffusion model to compute the noise vectors for testing samples because the testing samples need to be scaled corresponding to the in-distribution data points. To determine the scale factor of testing samples, we applied a binary search algorithm.
 
To summarize, in this paper:
\begin{enumerate}
    \item we introduce the diffusion model to tackle the out-of-distribution detection task. Our diffusion model-based OOD method largely outperforms the previous OOD method. The highest Auroc gain is above $3.5\%$ with an MAE-based image encoder.
    \item we analyze factors in the diffusion model that influence the accuracy of out-of-distribution detection. A closed-form noise vector formula is proposed to help us understand the noise vectors predicted by our deep diffusion model. Experiments on the OOD benchmark are conducted with both noise vectors predicted by the deep diffusion model and the closed-form noise vector. The two experiments closely agree with each other.
    \item We study the impact of an image encoder on out-of-distribution(OOD) detection, benchmark previous OOD methods, and compare the robustness of these methods with our diffusion method. Our diffusion-based method is robust with different types of image encoders and consistently outperforms previous OOD methods.
\end{enumerate}

\begin{figure*}[t!]
\centering
\includegraphics[width=1.0\textwidth]{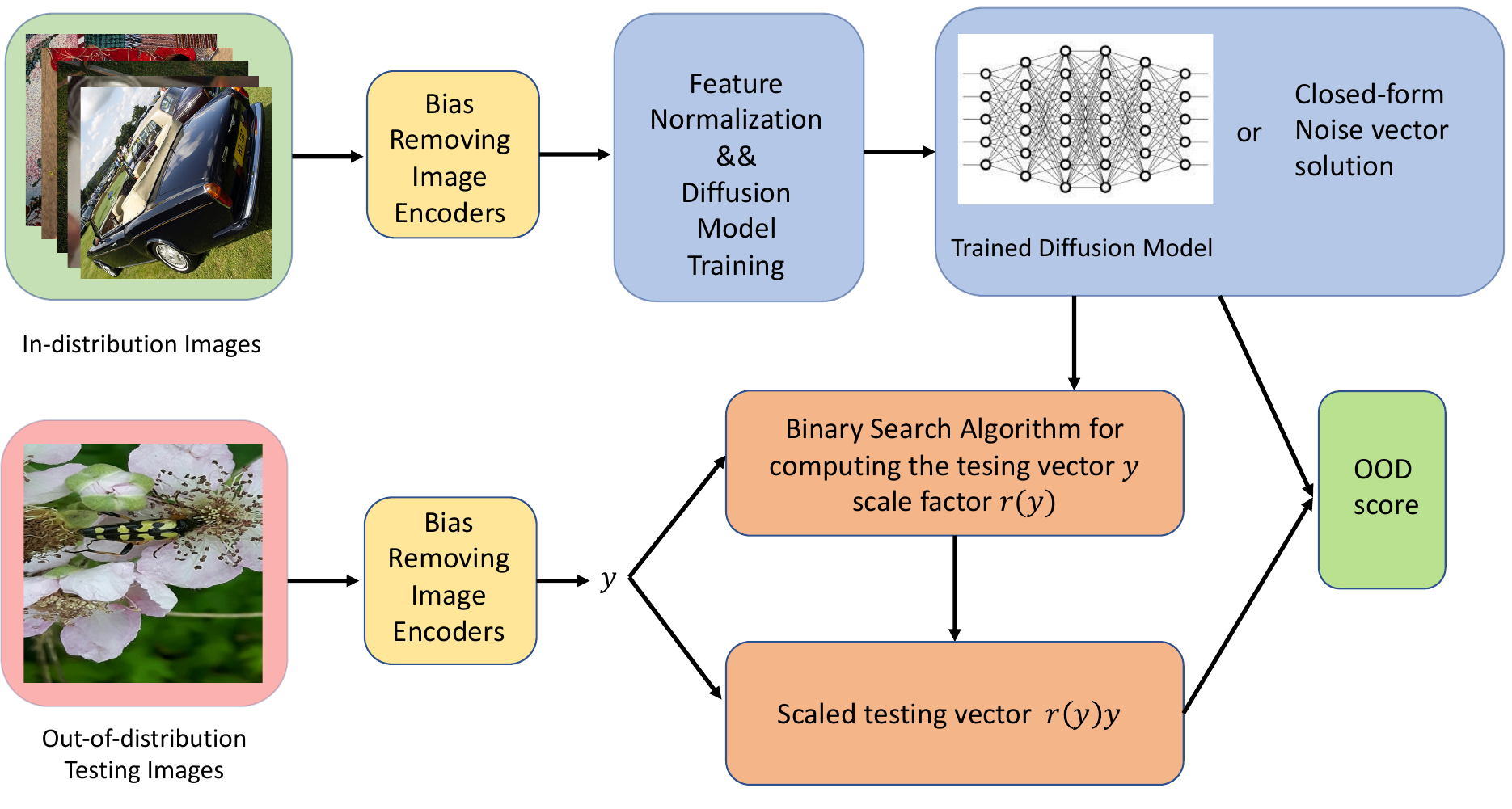}
\caption{The pipeline of Diffusion Based OOD. During the training of the diffusion model, the in-distribution images are first encoded with a bias removing image encoder. Then those in-distribution feature vectors are normalized and used to train a diffusion model or compute the closed-form noise feature vector. At test time, the testing images are encoded with the same bias removing image encoder. A binary search algorithm is applied to determine the scale factor $r(y)$ of the testing feature vector $y$. Together with the trained diffusion model or the closed-form noise vector formula, we compute the OOD score of the scaled testing feature vector $r(y)y$.}
\label{fig:framework}
\vspace{-3mm}
\end{figure*}

\section{Preliminaries}
\subsection{Problem Statement}
The Out-of-distribution (OOD) detection task aims to distinguish between in-distribution and out-of-distribution samples. In practice, An OOD score $\mathcal{S}(x)$ is assigned to each sample and a threshold $\gamma$ is used to determine whether a sample lies within in-distribution or out-of-distribution. However, the precision and recall of an OOD score function depend heavily on the choice of the OOD threshold $\gamma$. Thus the AUROC(Aera under the curve) metric is used to evaluate the effectiveness of a given OOD score function $\mathcal{S}(x)$. The key point of the OOD detection task is designing an OOD score function with high AUROC.
\subsection{Diffusion Model}
We provide a brief introduction to the diffusion model (DM). A diffusion model is a generative model that estimates the noise added to the training data. During training, a forward process of corrupting data is applied first. At each time step, the data point $x_t$ is updated as:
\begin{equation}\label{eq1}
x_t = \sqrt{{\alpha}_{t-1}} x_{t-1}+\sqrt{1-{\alpha}_{t-1}} \epsilon_{t-1}
\end{equation}
where $\alpha_t$ is a coefficient used to control the noise magnitude at each time step, and $\epsilon_t$ is a random vector with the same dimension as $x_{t-1}$. For most existing work, elements of $\epsilon_t$ satisfy the standard normal distribution $\mathcal{N}(0,1)$. Suppose that each element of $\epsilon_t$ is drawn from the standard Gaussian distribution, one can rewrite \ref{eq1} as:
\begin{equation}\label{eq2}
    x_t = \sqrt{\bar{{\alpha}}_{t-1}} x_{0}+\sqrt{1-\bar{{\alpha}}_{t-1}} \bar{\epsilon}_{t-1}
\end{equation}
where $\bar{\alpha}_{t}=\prod_{i=0}^{t}\alpha_i$ and $\bar{\epsilon}_t$ is a random vector with each element satisfying the standard normal distribution. Then a deep neural network $\mathcal{M}_{\theta}$ parameterized by $\theta$ is applied to estimate this Gaussian noise $\bar{\epsilon}_{t}$ given the time step $t$ and perturbed data point $x_t$.
\begin{equation}
    {\eta}_{t} =  \mathcal{M}_{\theta}(x_t,t)
\end{equation}
To obtain the parameter $\theta$, we minimize the following diffusion loss:
\begin{equation}\label{loss}
\mathcal{L}_{diff} = \mathop{\mathbb{E}}_{x_0,t}\mathop{\mathbb{E}}_{\bar{\epsilon}_t}({\eta}_t-\bar{\epsilon}_t)^2
\end{equation}

\section{Method}
In this section, we explain our method of introducing the diffusion model into out-of-distribution detection. As shown in Fig.~\ref{fig:framework}, we encode the in-distribution images with neural image encoders, this process requires a specially designed bias-removing mechanism. Then we normalized all the encoded bias-removing features onto the boundary of a closed ball set $\partial \mathcal{B}(0,r)=\{x|x\in\mathbb{R}^n,\|x\|_2=r\}$. These normalized image features are used to train a diffusion model and compute closed-form noise vectors. Finally, we estimate the noise component of each testing sample with our trained diffusion model and closed-form noise vector formula. During the testing process, each testing vector $y$ should be scaled by a scale $r(y)$, and this $r(y)$ is determined through binary search.
\subsection{Bias Removing Image Encoding}\label{encode}
In previous research \cite{msp,deepsvd,ctsood,vim}, images are usually encoded as features first, however little has been done to analyze the impact of different image encoders and the OOD methods' robustness across various image encoders. Thanks to the recent development in vision transformer and representation learning, we can extend the categories of image encoders from the ResNet-50 classification model to models of different architectures and training procedures. By conducting experiments on a variety of image encoders, we can justify the robustness of our method.  Aside from the ResNet-50 classification model, we also applied deep models like Beit \cite{bao2021beit}, Deit \cite{deit}, MAE \cite{MAE} as our image encoders.

Let $I$ be an image and $\mathcal{E}$ be a neural image encoder that encodes an image in the latent space $\mathbb{R}^d$. Suppose the classification head that following $\mathcal{E}$ is a linear layer with weight $W\in\mathbb{R}^{d\times C}$ and bias $b\in\mathbb{R}^C$, where $C$ is the number of class in the training set. Now we denote the latent vector $x_{I}$ for image $I$ and the probability $p(x_I)$ of of $I$ belonging to each class as:
\begin{align}\label{eq3}
    x_I &= \mathcal{E}(I) \nonumber \\
    p(x_I) &= \textrm{softmax}(W^Tx_I+b)
\end{align}
As discussed in \cite{vim}, the bias vector $b$ contributes to shift the distribution of $p(x_I)$ thus the encoded feature must address this distribution shifting information. We make the assumption that if we can find a vector $y_I$ such that:
\begin{align}\label{eq4}
    p(x_I) = {\rm softmax}(W^Ty_I)
\end{align}
Then this distribution shifting information caused by bias $b$ is integrated into $y_I$. The author of \cite{vim} proposed to use $y_I=x_I+(W^T)^{+}b$ as the encoded features, where $(W^T)^{+}$ is the Moore-Penrose inverse. However, \ref{eq4} can easily fail when we are encoding with vision transformers. Because $W^T$ is not of full column rank in vision transformers, this leads to the fact that $b$ may not be in the range of $W^T$ \cite{zhou2018distributed, zhou2019finite, wang2017distributed,wang2019distributed}.  

To address the above issues, we propose our bias-removing strategy. Columns are appended at the end of $W^T$ and form:
\begin{equation}
    \bar{W}^T = [W^T, P^T],
\end{equation}
where $P^T$ is a set of bases for the kernel of $W^T$. In this way, the columns of $\bar{W}^T$ span the whole linear space $\mathbb{R}^C$, and we brought $b$ back to the range of $\bar{W}^T$.  To maintain predicted probabilities, zeros are padded at the end of the latent vector $x_I$ to form:
\begin{equation}
\bar{x}_{I}^T=[x_{I}^T,0^T]\in\mathbb{R}^C
\end{equation}
Through the above constructions, we can claim that:
\begin{equation}\label{eq5}
    p(x_I) = {\rm softmax}(\bar{W}^T(\bar{x}_{I}+(\bar{W}^T)^{+}b))
\end{equation}
The image $I$ is encoded as:
\begin{equation}
    y_I=\bar{x}_I+(\bar{W}^T)^{+}b
\end{equation}
We encode all images on the in-distribution training set, and we denote $\mathcal{Y}_{c}$ as the encoded in-distribution training set for class-$c$.

\subsection{Normalized Diffusion and Stable Point}\label{Diffusion}
The out-of-distribution detection (OOD) task is essentially a binary classification task. However, the OOD task differs from the usual classification task in two major aspects. 1): Only in-distribution data is provided, thus common classification methods can not handle this situation; 2). The scope of out-of-distribution data is unlimited. Unlike the naive classification problem that each class contains only one category of objects, the out-of-distribution data is formed from an infinite number of categories of data that differ from the in-distribution data. For this reason, it is hard to craft a typical set of out-of-distribution data.

The diffusion model is able to address the above two issues. By adding noise $\epsilon_t$ to our in-distribution encoded feature $x_{id}$, we can create out-of-distribution data $x_{od}$. Conversely, during the testing stage, we can estimate the noise vector $\eta_t$ given the testing sample feature vector $y$, which can be further processed to produce an OOD score.

To use the diffusion process in our OOD task, we need to normalize the whole in-distribution dataset to a given magnitude $r$. The resulting normalized feature of class $c$ is:
\begin{equation}
    \mathcal{D}_{c} = \left\{r \frac{y}{\|y\|}| y\in\mathcal{Y}_c\right\}
\end{equation}
where $\mathcal{Y}_c$ is the encoded in-distribution feature set as described in ~\ref{encode}. Let $x_0\in \bigcup_{0\leq c < C}\mathcal{D}_{c}$ denote an in-distribution feature at the first diffusion step. Similarly, we have $x_t$ stand for the feature after $t$ steps of Gaussian diffusion. We can deduce an analysis solution to the noise vector $\eta_{t-1}(y)$ at time step $t$ for the testing vector $y$. We call this $\eta_{t-1}(y)$ as a \textbf{stable point} and it can be further processed to compute an OOD score for testing sample $y$. To be more rigorous, we rewrite \ref{loss} as:
\begin{align}\label{loss_rewrite}
    \mathcal{L}_{diff} &= \mathop{\mathbb{E}}_{t}\mathop{\mathbb{E}}_{x_0}\mathop{\mathbb{E}}_{\frac{x_t-\sqrt{\bar{\alpha}_{t-1}}x_0}{\sqrt{1-\bar{\alpha}_{t-1}}}|x_0,t}({\eta}_{t-1}-\frac{x_t-\sqrt{\bar{\alpha}_{t-1}}x_0}{\sqrt{1-\bar{\alpha}_{t-1}}})^2\nonumber \\&=\mathop{\mathbb{E}}_{t}\mathop{\mathbb{E}}_{x_0}\mathop{\mathbb{E}}_{x_t|x_0,t}\frac{1}{\sqrt{1-\Bar{\alpha}_{t-1}}}({\eta}_{t-1}-\frac{x_t-\sqrt{\bar{\alpha}_{t-1}}x_0}{\sqrt{1-\bar{\alpha}_{t-1}}})^2\nonumber \\&=\mathop{\mathbb{E}}_{t}\mathop{\mathbb{E}}_{x_t}\mathop{\mathbb{E}}_{x_0|x_t}\frac{1}{\sqrt{1-\Bar{\alpha}_{t-1}}}({\eta}_{t-1}-\frac{x_t-\sqrt{\bar{\alpha}_{t-1}}x_0}{\sqrt{1-\bar{\alpha}_{t-1}}})^2.
\end{align}
The last equation $x_0|x_t$ does not involve $t$ and comes from the fact that $x_0$ is independent of $t$. Due to \ref{loss_rewrite}, the loss $\mathcal{L}_{diff}$ is a convex function with respect to the predicted noise $\eta_{t-1}$ for each testing sample $x_t$. Then we have
\begin{align}\label{stable_point}
    0 & = \frac{\partial{\mathop{\mathbb{E}}_{x_0|x_t}\frac{1}{\sqrt{1-\Bar{\alpha}_{t-1}}}({\eta}_{t-1}-\frac{x_t-\sqrt{\bar{\alpha}_{t-1}}x_0}{\sqrt{1-\bar{\alpha}_{t-1}}})^2}}{\partial{\eta_t}}\nonumber\\
    & = \mathop{\mathbb{E}}_{x_0|x_t}\frac{1}{\sqrt{1-\Bar{\alpha}_{t-1}}}({\eta}_{t-1}-\frac{x_t-\sqrt{\bar{\alpha}_{t-1}}x_0}{\sqrt{1-\bar{\alpha}_{t-1}}}).
\end{align}
By \ref{stable_point}, we consider the case that $x_0 \in \mathcal{D}_c$ reaches $y$ after $t$ steps of diffusion. Then it follows to replace $x_t$ with $y$:
\begin{align}\label{stable}
    \eta_{t-1}(y,c) &= \frac{1}{\sum_{x_0}p(x_0|y)}\sum_{x_0}p(x_0|y)\frac{y-\sqrt{\bar{\alpha}_{t-1}}x_0}{\sqrt{1-\bar{\alpha}_{t-1}}}\nonumber\\
            & = \frac{1}{\sum_{x_0}p(x_0,y)}\sum_{x_0}p(x_0,y)\frac{y-\sqrt{\bar{\alpha}_{t-1}}x_0}{\sqrt{1-\bar{\alpha}_{t-1}}},
\end{align}
where $p(x_0,x_t)$ is the probability of sampling $x_0$ and $x_t$ at the same time. In the diffusion model, this probability is Gaussian. The stable point method is a direct utilization of the diffusion process in OOD task. Although it is time-consuming, it serves as an anchor point for the performance of deep-model-based diffusion OOD.

Suppose we are given a diffusion model $\mathcal{M}_{\theta}(y,t,c)$ with sufficiently large model capacity. Then it is possible to fit Eq. \ref{stable} with loss Eq. \ref{diff_loss}, and the diffusion model $\mathcal{M}_{\theta}(y,t,c)$ should also make reasonable prediction given the testing vector $y$ and diffusion time $t$ and specified class $c$. The advantage of using this deep-diffusion model is faster inference speed because it avoids computation on the whole training set. In this case, we predict the noise vector as:
\begin{equation}
    \eta_{t-1}(y,c) = \mathcal{M}_{\theta}(y,t,c)
\end{equation}

\subsection{Noise Vector-Based OOD Score}
\begin{figure*}[t!]
\noindent\begin{minipage}{.49\textwidth}
\centering
\noindent\begin{algorithm}[H]
\caption{Binary Search for $r(y)$}\label{ry}
\label{alg:depth_width}
\small{
	\begin{algorithmic}[1]
	    \State$r_{low}=\frac{r}{2}$ $r_{high}=2r$
            \State $err =1$

            \While{$err >thr$}
            \State $r(y) = \frac{r_{low}+r_{high}}{2}$
            \State $ r_t= \|\frac{r(y)y-\sqrt{1-\bar{\alpha}_{t-1}}\eta_{t-1}(r(y)y,c)}{\sqrt{\bar{\alpha}}_{t-1}}\|$
            \State $err = |r_t-r|$
            \If{$r_t>r$}
                \State $r_{low} = r(y)$
            \ElsIf{$r_t\leq r$}
            \State $r_{high} = r(y)$
            \EndIf
            \EndWhile
	\end{algorithmic}
    }
\end{algorithm}
\hrule
\end{minipage}
\begin{minipage}{.49\textwidth}
    \centering
    \includegraphics[width=0.8\linewidth]{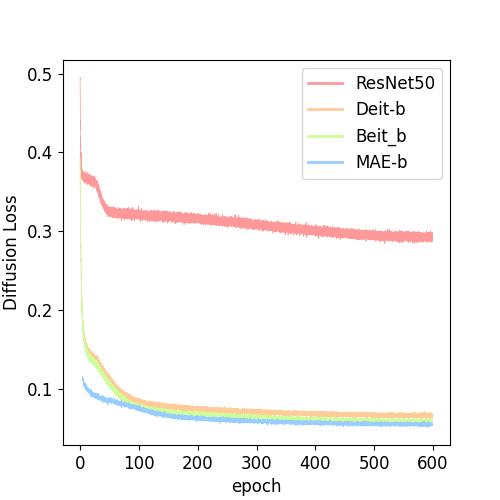}
    \captionof{figure}{Diffusion loss for diffusion model under different image encoders.}
  \label{diff_loss}
\end{minipage}%
\vspace{-5mm}
\end{figure*}

In sections \ref{encode} and \ref{Diffusion}, we deduced the formula of stable point and trained a diffusion model that can predict the noise vector between a testing sample and the in-distribution dataset. During the diffusion process, the in-distribution vectors are all normalized to a constant magnitude $r$. The normalization procedure in section \ref{Diffusion} implicitly indicates that all in-distribution data are on the surface of the close ball $\mathcal{B}(0,r)=\{x|x\in\mathbb{R}^n,\|x\|_2\leq r\}$. Then it is natural to assume that, all testing feature $y$ must be originated on some vector $x_0\in \{x|x\in\mathbb{R}^n,\|x\|_2=r \}$ during the diffusion process. To enforce this assumption, we scale the length of the testing sample feature vector by a factor $r(y)$, as:
\begin{equation}
   \|x_0\|= \left\|\frac{r(y)y-\sqrt{1-\bar{\alpha}_{t-1}}\eta_{t-1}(r(y)y,c)}{\sqrt{\bar{\alpha}}_{t-1}}\right\|=r
\end{equation}
where $\eta_{t-1}(r(y)y,c)$ can be predicted by the stable point formula or the trained deep diffusion model. It's non-trivial to find this $r(y)$ since the non-linearity in $\eta_{t-1}(r(y)y,c)$. Then, we applied the binary search algorithm (Algorithm~\ref{ry}) to determine the value of $r(y)$. In this algorithm, $thr$ is a predefined threshold for iteration termination.

Note that the process to determine the value of $r(y)$ depends on the selection of diffusion time-step $t$. Here $t$ can be viewed as a hyperparameter and we choose $t=3$ for the ResNet50 encoded model and $t=2$ for all Vision Transformer encoded models. For each possible class $0\leq c<C$, we compute a noise vector $\eta_{t-1}(r(y)y,c)$. The smallest magnitude of all these noise vectors is chosen to be the OOD score. That is the OOD score $\mathcal{S}(y)$ for testing sample $y$ is:.
\begin{equation}
    \mathcal{S}(y) = \min\{{\|\eta_{t-1}(r(y)y,c)\|} \;| 0\leq c <C\}
\end{equation}

\section{Experiments}
In this section, we evaluate the diffusion-based OOD method on various image encoders, also, these image encoders are used to benchmark previous OOD detection algorithms. The experimental setup is described in Section \ref{setup}, and we exhibit the overall 
superior performance of diffusion-based OOD over existing approaches in Section \ref{result}. Then we conducted extensive ablation studies for understanding this approach.
\subsection{Experimental Setup}\label{setup}
\textbf{Diffusion Model:} We use a neural network with multiple residual blocks as our diffusion deep model. During the training process, the diffusion max time step is set to be $10$, and $\alpha_t$ increases linearly from $10^{-4}$ to $10^{-2}$. The diffusion model is initialized with random weight uniformly sampled from $(0,1)$ and trained with a cos learning rate schedule whose $lr_{high}=0.01$ $lr_{low}=0.0001$. The whole training process takes 600 epochs.

\textbf{Stable Point:} The stable point formula is given by \ref{stable}, and the prediction of the deep models should converge to these stable points with enough model capacity. Due to the Gaussian sampling strategy in the diffusion model, the stable point $\eta_{t-1}(r(y)y,c)$ is computed as:
\begin{equation}
    \eta_{t-1}(r(y)y,c)=\frac{1}{\sum_{x_0}exp(-\frac{\epsilon_{t}^{T}(x_0)\epsilon_{t}(x_0)}{2})}\sum_{x_0}exp(-\frac{\epsilon_{t}^{T}(x_0)\epsilon_{t}(x_0)}{2})\epsilon_{t}(x_0)
\end{equation}
where $\epsilon_{t}(x_0)=\frac{r(y)y-\sqrt{\bar{\alpha}_{t-1}}x_0}{\sqrt{1-\bar{\alpha}_{t-1}}}$ and $r(y)$ is determined with binary search as in Algorithm \ref{ry}. 

\textbf{Dataset:} In this work, we focus our experiments and ablation studies on the large-scale dataset Imagenet.  Following the settings in 
\cite{ctsood}, OOD datasets are categorized as far-ood datasets and near-ood datasets. For the near-ood datset, it contains: Species \cite{species}, iNaturalist \cite{inautralist}, ImageNet-O \cite{imageneto}, OpenImage-O \cite{vim}. The far-ood dataset consists of Texture \cite{texture}, Mnist \cite{mnist}. ImageNet is used as our in-domain dataset, the diffusion model is trained on the training split and tested on the validation split.

\subsection{Results and Ablations}\label{result}
\textbf{Comparison with Existing OOD Methods:} Our main result is shown in Tab. \ref{main}, throughout these experiments, we set $r=7$ for ResNet50 and $r=4$ for Vision Transformers. On the near-ood split, our diffusion model surpasses the previous method from $3.7\%$ (resnet50) to $2.34\%$(Beit-b) in terms of AUROC. On the relatively easier far-ood split, our diffusion model achieves $1\%$ higher Auroc on all vit-based image encoders. Then we conclude that our diffusion-based OOD method largely outperforms previous OOD methods on all kinds of image encoders.

Besides, we find that our diffusion-based OOD model has a consistent performance with its \textbf{stable point} counterpart. On all three Vit-based image encoders, the diffusion models achieve Auroc almost identical to the stable point method. However, the ResNet50 encoded diffusion model performance is worse than the stable point result. We notice that the feature dimension of ResNet50 is $2048$ and the Vit-base is $768$, thus it's more difficult to fully sample the Gaussian noise space and requires a larger network capacity to fit the noise vector.

To support our claim,  We design the following experiments and compare the training loss between diffusion models with different kinds of image encoders. The normalization scale is $r=4$ and other experiment settings are as stated in section \ref{setup}.  As in Fig.~\ref{diff_loss}, the losses of diffusion models encoded by Vision transformers are only $20\%$ that of the diffusion model encoded by ResNet50. This suggests the ResNet50-encoded diffusion model converges more poorly than the Vision Transformer-encoded ones. Furthermore, we computed the theoretical mean square error for the noise vector of the ResNet50-encoded diffusion model, the value is $\mathcal{L}_{theory}=0.056$ which is far less the value shown in \ref{diff_loss}. For the above reason, we conclude the large performance gap between the diffusion model and the stable point method in the ResNet50 case is caused by training difficulty and network capacity.


Lastly, we investigated the robustness of different OOD methods on different types of image encoders. Generally speaking, we find out that OOD methods utilize the classification score \cite{mls,msp} or feature distance \cite{sun2022out,vim} are robust over different image encoders.  However, the method that utilizes the gradient information \cite{odin} fails to generalize well when provided with a vision transformer encoder rather than ResNet. What's more our diffusion-based OOD method exhibits good robustness to image encoders on the OOD benchmark data set. Both the deep diffusion model version and the stable point version achieve high performance on the OOD benchmark with various image encoders.

\begin{table}[!t]
\begin{center}
\caption{\textbf{OOD detection for NOID and state-of-the-art methods}}
\label{main}

\begin{tabular}{c |c |c|c c|c c}
\hline
\hline
Method name &encoder&Acc& Near Fpr@95&Near Auroc&Far@Far Fpr@95&Far Auroc \\
\hline
\hline
MSP \cite{msp} & \multirow{7}{*}{ResNet50} &  \multirow{7}{*}{76.1} &  74.15 &  69.26 & 58.26& 86.11\\
ODIN \cite{odin} &  &   &  68.34 &  73.07 & 21.62& 94.39\\
DICE \cite{dice} &  &   &  65.22 &  73.62 & 15.84& 95.67\\
MLS \cite{mls} &  &   &  72.39 &  73.47 & 37.97 &92.23\\
KNN \cite{sun2022out} &  &   &  71.26 &  81.03 & 8.43& 98.04\\
VIM \cite{vim} &  &   &  73.34 &  79.99 & 6.86& \textbf{98.37}\\
NODI(ours) &  &   &  62.49 &  84.80 & 38.22& 95.48\\
Stable Point(ours)&  &   &  33.56 &  \textbf{93.22} & 20.31& 95.62\\
\hline
\hline
MSP \cite{msp} & \multirow{7}{*}{Deit-B} &  \multirow{7}{*}{81.8} & 69.79&77.44&59.36&85.61\\
ODIN \cite{odin} &  &   & 77.27	&64.50&70.95&64.27\\
DICE \cite{dice} &  &   &  88.96&52.44&39.2&88.37\\
MLS \cite{mls} &  &   &  69.87	&74.20&56.76&83.08\\
KNN \cite{sun2022out} &  &   & 81.31&81.71&74.34&89.78\\
VIM \cite{vim} &  &   &  74.62&81.93&80.73&87.61\\
NODI(ours) &  &   &  61.77 &  84.64 & 54.36& 90.81\\
Stable Point(ours)&  &   &  61.34 &  \textbf{85.1} & 54.28& \textbf{90.82}\\
\hline
\hline
MSP \cite{msp} & \multirow{7}{*}{Beit-B} &  \multirow{7}{*}{83.6} & 64.6&78.95&46.24&88.43\\
ODIN \cite{odin} &  &   & 83.01	&57.79&79.16&57.71\\
DICE \cite{dice} &  &   & 89.23&55.42&41.36&87.36\\
MLS \cite{mls} &  &   & 64.00&76.74&42.66&87.35\\
KNN \cite{sun2022out} &  &   &69.57&81.05&30.19&93.63\\
VIM \cite{vim} &  &   & 58.56&83.95&59.75&91.91\\
NODI(ours) &  &   &  51.94 &  86.29 & 36.67& 93.49\\
Stable Point(ours)&  &   &  52.76 &  \textbf{86.58} & 31.28& \textbf{93.93}\\
\hline
\hline
MSP \cite{msp} & \multirow{7}{*}{MAE-B} &  \multirow{7}{*}{83.6} &  65.23&79.18&63.33&83.10\\
ODIN \cite{odin} &  &   & 85.73&53.69&83.59&49.94\\
DICE \cite{dice} &  &   & 86.82&53.77&59.72&84.94\\
MLS \cite{mls} &  &   &  68.29&77.72&65.33&81.00\\
KNN \cite{sun2022out} &  &   & 74.22&83.47&45.76&92.93\\
VIM \cite{vim} &  &   & 59.37&84.81&43.92&93.69\\
NODI(ours) &  &   &  45.59 &  \textbf{88.31} & 19.27& \textbf{95.32}\\
Stable Point(ours)&  &   &  54.75 &  87.61 & 27.41& 94.23\\
\hline
\hline

\end{tabular}
\vspace{-5mm}
\end{center}
\end{table}
\textbf{Effect of Bias Removal and Feature Normalization}: In this section, we study the necessity of both removing bias vector $b$ of the prediction head as in section \ref{encode} and normalizing the feature vector as in section \ref{Diffusion}. As demonstrated in Tab. ~\ref{rmbia_norm}, both the bias-removing technique and normalization operation positively contribute to our deep-diffusion OOD model. The performance gain from removing bias supports our claim that integrating the classification bias $b$ into encoded features improves OOD performance.
\begin{table}[h!]
    \centering
    \begin{tabular}{c|c|c|c|c}
    \hline
    \hline
        Diffusion Model &  Fpr@95 (near) & Auroc (near) & Fpr@95 (far) & Auroc (far)\\
        \hline
         \hline
         DDPM & 79.34& 82.15&82.22 &89.32\\
                 \hline
        DDPM+RmBias & 68.20& 85.00& 24.71&94.55\\
         \hline
         DDPM+$r$-Norm & 62.29&84.60 &24.11 &94.71\\
          \hline
          DDPM+RmBias+$r$-Norm & 45.59& 88.31&19.27 &95.32\\
                 \hline
         \hline
    \end{tabular}
    \caption{Effect of bias removing and feature normalization. The abbreviation 'DDPM' stands for the original diffusion model \cite{ddpm}, 'RmBias' stands for the bias-removing procedure mentioned in section \ref{encode}, and '$r$-Norm' stands for the normalization process in section \ref{Diffusion}. All these experiments are done under the MAE-image encoder, $r$ is set to $4$, and other settings are the same as in section \ref{setup}.}
    \label{rmbia_norm}
    \vspace{-5mm}
\end{table}

\textbf{Normalization Factor}: In section \ref{Diffusion} we normalized the encoded image features to a predefined length $r$. This length $r$ determines the distance between in-distribution samples and influences the performance of our diffusion model. To study the impact of $r$, we conduct experiments on diffusion models encoded by the vision transformer image encoders (Deit, Beit, MAE). These experiments share the same setup as described in section \ref{setup}, except that the normalization factor $r$ is varied.  

\begin{figure}[h]
  \includegraphics[width=15cm,height=4.5cm]{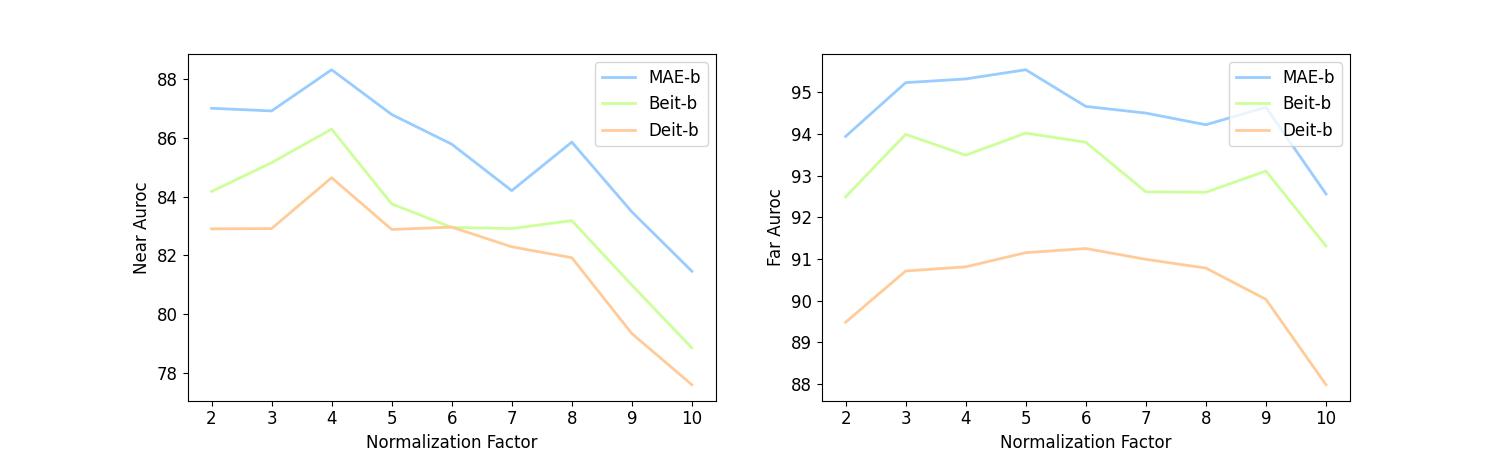}
  \caption{The impact of normalization factor $r$ on different types of image encoder.}
  \label{r_ablation}
  \vspace{-5mm}
\end{figure}

As shown in Fig.~\ref{r_ablation}, all the diffusion models achieve the highest Auroc when the diffusion normalization factor $r$ is around $4$. The diffusion model performance drops when $r$ takes a considerably large or small value. We claim that different reasons cause this phenomenon. 

For significantly large $r$, the diffusion model fails to capture the whole latent space because most noise vectors generated during the diffusion process can be viewed as slightly perturbed vectors of the in-distribution data points. The diffusion model tends to make random predictions when out-of-distribution data are provided, and this leads to a performance drop when the normalization factor is too large.

For very small $r$, although the diffusion model captures most of the latent space, it wastes too much model capacity in fitting noise vectors that are far from the in-distribution data points. For this reason, the diffusion model suffers some performance drop.

\textbf{Optimal Diffusion Time}: As shown in Tab.~\ref{t_impact}, the diffusion time $t$ negatively impacts the performance of our diffusion model when $t=0$. This is similar to the case when the normalization length $r$ is too large. The out-of-distribution samples can not be sampled enough when diffusion time $t$ is small. Different from the normalization factor $r$, the performance does not drop significantly low when $t$ reaches extremely large values.
\begin{table}[h!]
    \centering
    \begin{tabular}{c|c|c|c|c|c|c|c|c|c|c}
    \hline
    \hline
        Diffusion Time & 0 &1& 2 & 3 &4&5&6&7&8&9\\
        \hline
         \hline
         Near Auroc &85.79&88.42&88.31&88.43&88.42&88,38&88.32&88.26&87.86&87.60\\
                 \hline
        Far Auroc &94.26&95.64&95.32&95.62&95.67&95.66&95.65&95.66&95.03&95.02\\

          \hline
         \hline
    \end{tabular}
    \caption{Impact of diffusion time-step $t$ on the Auroc performance}
    \label{t_impact}
    \vspace{-5mm}
\end{table}

\begin{figure}[h]
  \includegraphics[width=15cm,height=4.5cm]{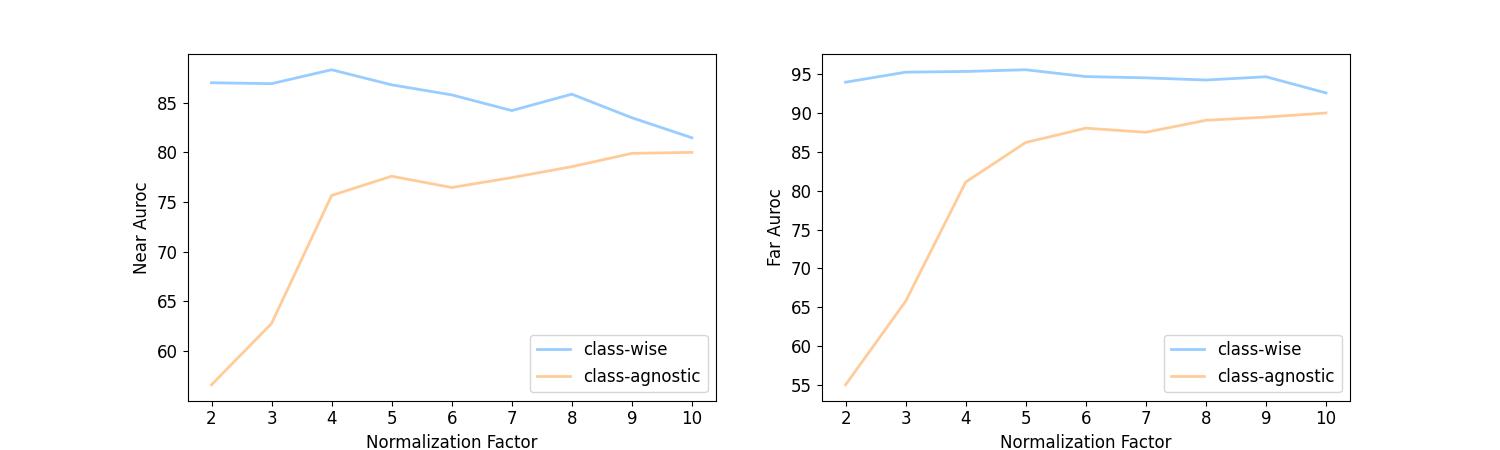}
  \caption{Class-wise OOD score and Class agnostic OOD score}
  \label{class_ablation}
  \vspace{-5mm}
\end{figure}

\textbf{Class agnostic OOD and Class-wise OOD}: In our diffusion model, we compute the OOD score class-wisely and take the minimal one. It's natural to ask if this class-wise computation really necessary. To support our class-wise OOD score, we conduct experiments on the MAE-encoded features. As demonstrated in Fig. \ref{class_ablation}, treating all the in-distribution data as one class significantly reduces the performance of our diffusion-based OOD method. Also, we observe that the performance gap between the class-wise OOD method and the class-agnostic OOD method reduces as the normalization factor $r$ enlarges. This is caused by insufficient diffusion of the in-distribution sample. When $r$ is large enough, those two methods are equivalent because each testing sample is diffused from the closest in-distribution data point.

\section{Conclusion}
In this paper, we introduce the diffusion process into the out-of-distribution task. We conduct experiments with both the closed-form solution of noise vector and the deep model predicted noise vector on  OOD benchmarks. Our method showed significant performance improvement compared and robustness with respect to different image encoders. 

{\small
\bibliographystyle{ieee_fullname}
\bibliography{egbib}
}

\end{document}